\documentclass{article}
\usepackage{style,times}
\usepackage{booktabs}
\usepackage{caption}
\usepackage{array}
\usepackage{graphicx}
\usepackage{array}
\usepackage{booktabs}
\usepackage{adjustbox}
\usepackage{amsmath, amsthm, amssymb}

\usepackage{hyperref}
\usepackage{url}
\begin{document}
\begin{center}
\title{Parameter-Efficient Transformer Embedding}
\end{center}
\author{Henry Ndubuaku \\ 
Cactus Compute, Inc. \\
London, United Kingdom \\
\texttt{henry@cactuscompute.com} \\
\And
Mouad Talhi \\
Department of Computing, Imperial College \\
London, United Kingdom \\
\texttt{mt924@ic.ac.uk} \\
}
\maketitle
\begin{abstract}
Embedding layers in transformer-based NLP models typically account for the largest share of model parameters, scaling with vocabulary size but not yielding performance gains proportional to scale. We propose an alternative approach in which token embedding vectors are first generated deterministically, directly from the token IDs using a Fourier expansion of their normalized values, followed by a lightweight multilayer perceptron (MLP) that captures higher-order interactions. We train standard transformers and our architecture on natural language inference tasks (SNLI and MNLI), and evaluate zero-shot performance on sentence textual similarity (STS-B). Our results demonstrate that the proposed method achieves competitive performance using significantly fewer parameters, trains faster, and operates effectively without the need for dropout. This proof-of-concept study highlights the potential for scalable, memory-efficient language models and motivates further large-scale experimentation based on our findings. The code for reproducing and pre-trained weights are available at \url{https://github.com/HMUNACHI/pete}.
\end{abstract}
\section{Introduction and Related Work}
Embedding layers, which are typically represented as 2D matrices of dimensions \(V \times d\) (where \(V\) is the vocabulary size and \(d\) the embedding dimension), often account for more parameters in linguistic transformer models than other layers. Despite this, they do not necessarily yield proportional performance gains~\cite{lan2019albert, rajbhandari2020zero, shen2020qbert}. Several factors are thought to contribute to this inefficiency. 

Firstly, sparsity issues can lead to under-optimized embeddings since rare tokens remain underrepresented during training~\cite{svenstrup2017hash}. More importantly, embeddings may induce redundancy by assigning dense vectors to tokens with overlapping semantic roles, thereby wasting capacity~\cite{lan2019albert}. Additionally, traditional embeddings do not exploit entropy-driven compression; they allocate excessive parameters to frequent tokens without addressing the inherent redundancy in token distributions~\cite{shu2017compressing}.

Numerous research efforts have aimed to alleviate this parameter inefficiency by employing compression, adaptive designs, and alternative representations. Cai et al. \cite{cai2023dimensionlifting} highlighted the inefficiency of high-dimensional embedding representations in knowledge graph embeddings, proposing the Dimension Lifting Network (LiftNet). LiftNet replaces wide embeddings with a narrow embedding layer followed by a dimension lifting network, though this approach is not directly generalizable to all use cases. Xu et al. \cite{xu2023tensorgpt} introduced TensorGPT, which leverages Tensor-Train Decomposition to reduce the embedding layer's parameter count by up to 38.4 times with minimal performance degradation. In TensorGPT, a pre-trained embedding is projected to a lower dimension to facilitate deployment, even though a large embedding table must still be initially trained. Yan et al. \cite{yan2021adaptivemaskedtwins} proposed an Adaptively-Masked Twins-based Layer that dynamically adjusts embedding dimensions based on feature values, achieving significant memory savings and improved parameter utilization, albeit with a more complex training regime. Wang et al. \cite{wang2020structured} presented Structured Embedding Compression, which uses matrix factorization and product quantization to reduce the parameter count. Similarly, Lan et al. \cite{lan2019albert} developed ALBERT, a lightweight model that ties embeddings with the transformer's hidden representations to reduce redundancy. Shen et al. \cite{shen2020qbert} introduced Q-BERT, which employs Hessian-based quantization for embedding layers, though this method is compute-intensive. Additionally, hash-based embeddings have shown promise; for instance, Svenstrup et al. \cite{svenstrup2017hash} replaced traditional embedding tables with hash embeddings, achieving substantial parameter reductions for large vocabularies, albeit with specialized training procedures. Furthermore, although various forms of weight tying (such as embedding-output weight tying) are used in state-of-the-art models for language generation, such techniques do not readily benefit classification and language understanding tasks where output layers typically do not match the dimensions of the embedding layers. Also, while sub-word tokenization limits vocabulary size explosion, we instead propose a solution where the vocabulary size is not an algorithmic complexity.

Despite these advances, further exploration is required to develop universally efficient and effective embedding mechanisms. Many existing approaches focus on compressing or factorizing a large, pre-existing embedding table (e.g., ALBERT \cite{lan2019albert}, TensorGPT \cite{xu2023tensorgpt}, Q-BERT \cite{shen2020qbert}), or use techniques like hashing \cite{svenstrup2017hash} which break any inherent ordering in token IDs. In this work, we take a different approach motivated by two observations: first, that embedding layers are primarily lookups (factorized projections with no token inter-dependence), with complex semantic and pragmatic relationships learned later in attention blocks; and second, that Byte-Pair Encoding (BPE) often assigns IDs based on frequency, embedding statistical information into the ID sequence which carries distinguishing information. We hypothesize that this statistical structure, while not perfectly semantic, can be leveraged. We propose to replace the \(V \times d\) lookup table by viewing the embedding as a function mapping normalized token IDs to vectors, \(f: [-1, 1] \to \mathbb{R}^d\), and approximating this function using a combination of a fixed analytical basis and a learnable component. Specifically, we generate a base representation deterministically from the normalized token ID using Fourier expansion, and then refine this representation with a shared, lightweight learnable network (MLP). The core novelty lies in this structured approximation, using a deterministic, ID-derived analytical mapping (Fourier basis) combined with a small learnable residual component, aiming to avoid the need to learn or store \(V\) distinct vectors.

Our first motivation is that embedding layers in transformer models are fully factorized projections with no token inter-dependence; the semantic and pragmatic relationships are learned in the attention blocks. Our second motivation is that Byte-Pair Encoding (BPE) assigns token IDs based on each token's frequency in the corpus, effectively following a statistical pattern. The token IDs carry distinguishing information. To this end, we argue that legacy embedding layers could be approximated by first transforming discrete token IDs into continuous values in the range \([-1, 1]\) (described in the next section), computing their Fourier expansions up to a predetermined degree \(n\) which corresponds to the desired embedding dimension, then projecting onto a more aligned vector space with a shared Multi-Layered Perceptron. Other polynomial bases (e.g., Chebyshev, Legendre, Taylor) could be employed, however Taylor polynomials require the computation of derivatives, Chebyshev polynomials exhibit auto-regressive properties, and Legendre polynomials involve factorial computations. These make them less amenable to optimizations on accelerators such as GPUs, TPUs and NPUs. Computing Fourier on the fly at inference is still computationally expensive compared to merely mapping token IDs to learned vectors, however they are very amenable to hand-crafted hardware-aware implementations, as we have done in this work. 
\section{Methodology}
We adopt a parameter-efficient strategy to encode tokens by leveraging fixed Fourier basis functions combined with a learnable multilayer perceptron (MLP). This design is motivated by an information-theoretic perspective. In particular, BPE tokenization splits the input text into statistically significant word/subword units that are \emph{entropy-efficient} in terms of compression~\cite{sennrich2016neural}. Recall that for a token \(t\) with probability \(p(t)\), the \emph{surprisal} (self-information) is given by
\[
I(t) = -\log p(t),
\]
so that rare tokens (with low \(p(t)\)) contribute higher information. In many BPE schemes, token IDs are assigned in frequency order, meaning that frequent tokens receive lower IDs while rare tokens are assigned higher IDs. Although the token ID itself is an arbitrary label, its ordering reflects statistical properties of the vocabulary.
To harness this structure, we first normalize a token's integer ID \(p\) into the continuous interval \([-1,1]\) by defining
\[
x \;=\; 2\,\frac{p}{\texttt{vocabulary\_size}-1} \;-\; 1.
\]
This normalization maps discrete token IDs into a continuous, scale-invariant domain, allowing subsequent smooth transformations. Importantly, while the mapping is deterministic, it preserves the relative differences among tokens so that minor alterations (e.g., a word swap) affect the vector's magnitude more than its overall direction.
Next, we expand \(x\) into a high-dimensional embedding using a Fourier basis. For a chosen model dimension \(d_{\mathrm{model}}\), the token embedding \(\mathbf{T}(p) \in \mathbb{R}^{d_{\mathrm{model}}}\) is defined component-wise as
\[
T_{i}(p) 
\;=\;
\begin{cases}
\sin\Bigl(\bigl(\lfloor i/2 \rfloor + 1\bigr)\,\pi\,x\Bigr), & \text{if } i \text{ is even},\\[6pt]
\cos\Bigl(\bigl(\lfloor i/2 \rfloor + 1\bigr)\,\pi\,x\Bigr), & \text{if } i \text{ is odd}.
\end{cases}
\]
The choice of the Fourier basis is motivated by several factors from a function approximation perspective. Fourier series provide a well-known universal basis for approximating functions on a bounded interval like \([-1, 1]\), meaning that any sufficiently smooth target embedding function could theoretically be represented given enough basis components. The sine and cosine components form an orthogonal basis with respect to the standard L2 inner product over the continuous interval, which can help in creating less correlated features initially. Furthermore, the basis functions have a natural frequency interpretation: lower-order terms (\(\lfloor i/2 \rfloor\)) capture low-frequency variations across the normalized ID space (potentially corresponding to broad statistical trends), while higher-order terms capture finer details. Computationally, these functions are also amenable to efficient hardware implementation via custom kernels. It is crucial to note, however, that the effectiveness of using this fixed basis for the initial representation relies entirely on the empirical hypothesis that the statistically-driven ordering of BPE token IDs contains exploitable structure that correlates, even weakly, with the desired final embeddings.
Here, lower-order Fourier terms (e.g., \(T_0(x)\) and \(T_1(x)\)) capture global, coarse-grained information, while higher-order terms (\(T_n(x)\) for \(n\geq2\)) encode finer details. Under the small-angle approximation, the difference between embeddings of adjacent tokens (i.e., \(p\) and \(p+1\)) is approximately
\[
\Delta x \;=\; \frac{2}{V-1},
\]
so that with a large vocabulary \(V\), adjacent token embeddings in raw Fourier space lie very close. Moreover, because BPE token assignment is not strictly semantically monotonic (e.g., the tokens "cat" and "cathedral" might receive consecutive IDs despite semantic differences), the Fourier expansion alone may not sufficiently differentiate tokens with similar IDs.
To mitigate this, we append a learnable MLP to the Fourier features. The final token representation is given by
\[
\mathbf{E}(p) \;=\; \text{MLP}\bigl(\mathbf{T}(p)\bigr) + \mathbf{T}(p).
\]
The MLP operates on the fixed Fourier features, and the final embedding incorporates a residual connection. This architecture allows the MLP to learn a non-linear transformation that *corrects* or *refines* the initial deterministic representation \(\mathbf{T}(p)\). Its role is crucial for adapting the generic basis expansion to the specific nuances required by downstream tasks and for separating tokens whose base Fourier representations might be too similar. The residual connection explicitly frames the learned component as the adjustment needed relative to the fixed basis, potentially easing optimization by making it easier to learn small modifications or even the identity transformation if the base features are already effective.
From the universal approximation viewpoint, given any continuous target embedding function \(f: [-1,1] \to \mathbb{R}\), the MLP can approximate the residual function \(H(z)=f\bigl(\varphi^{-1}(z)\bigr)-z\) (where \(\varphi(x)=\mathbf{T}(p)\)) uniformly. That is, for every \(\epsilon>0\) there exists an MLP \(M\) such that
\[
\sup_{x \in [-1, 1]} \Bigl|\, M\bigl(\varphi(x)\bigr) + \varphi(x) - f(x) \Bigr| < \epsilon.
\]
Finally, although dropout is commonly used to reduce overfitting, we observed that applying dropout to these normalized continuous mappings disrupts the smooth progression of token IDs and degrades performance as expected in Polynomial-based approximations. The continuous normalization itself appears to provide a regularizing effect, contributing to the minimal overfitting observed even in over-parameterized regimes.
\section{Experimentation and Results}
Due to resource constraints including compute (a single Nvidia RTX 4090), team size, and time limitations, our experiments are intentionally scaled down to a proof-of-concept design, and are not optimized for main tracks at top conferences. Different neural architectures require different optimal hyper-parameters, but we evaluate the proposed transformer with Fourier embeddings (denoted as Fourier embeddings henceforth) and the corresponding baseline transformer under identical settings (established setups for transformers), differing only in the embedding layer. In the baseline transformer, the token embeddings are learned conventionally. Although extensive pre-training and evaluation at much larger model sizes, on more benchmarks (e.g., GLUE or specialized reasoning tasks) would provide deeper insights, we defer these directions to future work. Our focus here is to assess how effectively an attention-based model can learn semantic information using a semi-approximated embedding mechanism.

Our experimental setup employs mixed-precision training on the entailment subsets of the SNLI and MNLI datasets~\cite{snli,mnli} with a batch size of 128 over 122,700 iterations, a learning rate of \(2\times10^{-5}\), and 1,000 warmup steps. For the baseline transformer, we use a dropout probability of 0.1 (yielded best results), whereas Fourier embedding omits dropout. Fourier embedding implements its embedding mechanism via a custom CUDA kernel that fuses normalization and Fourier expansion. We utilize the BERT Tokenizer~\cite{devlin2019bert} (vocabulary size 30,522), rotary positional encoding~\cite{su2021rotary}, root-mean-squared layer normalization~\cite{liu2020rmsnorm}, and GeGLU activation~\cite{shazeer2020glu}. In addition, we employ average pooling and the AdamW optimizer~\cite{loshchilov2019decoupled}. Following the Fourier expansion, we include a position-wise feed-forward block with an intermediate expansion factor of 4 and GeGLU activation. While a large MLP might nearly match the parameter count of a learned embedding matrix, our experiments indicate that replacing this block with a simple linear layer only marginally degrades performance, while further reducing model size. Our primary objective is to isolate the impact of substituting the learned embedding layer with a deterministic alternative.

For training, we employ a contrastive loss function inspired by CLIP~\cite{radford2021learning} and InfoNCE~\cite{oord2018representation}. This loss encourages embeddings corresponding to matching sentence pairs (or entailment pairs) to be close in the embedding space, while non-matching pairs are pushed apart. A learnable temperature parameter is used to appropriately scale the cosine similarity scores. This contrastive framework leverages the entailment data (approximately 200k samples) to enforce semantic consistency in the learned representations.
Table~\ref{tab:model_scores} summarizes our main findings.
\begin{table}[h!]
    \centering
    \begin{adjustbox}{max width=\textwidth}
    \begin{tabular}{|c|c|c|c|c|c|c|}
        \hline
        \textbf{Model} & \textbf{Layers/Heads} & \textbf{d-model} & \textbf{Params} & \textbf{STSB Spearman-R} & \textbf{STSB Pearson-R} & \textbf{Training Time}\\
        \hline
        Transformer (Fourier Embedding) & 1 & 256 & 1.1m & 74.93 & 74.54 & 37.88 min\\
        Transformer & 1 & 256 & 8.9m & 77.01 & 76.80 & 48.48 min\\
        \hline
        Transformer (Fourier Embedding) & 1 & 512 & 4.7m & 75.21 & 74.65 & 1.349 hr\\
        Transformer & 1 & 512 & 20.1m & 77.50 & 76.78 & 1.688 hr\\
        \hline
        Transformer (Fourier Embedding) & 2 & 256 & 2.2m & 76.38 & 76.02 & 1.009 hr\\
        Transformer & 2 & 256 & 9.9m & 77.34 & 76.89 & 1.322 hr\\
        \hline
        Transformer (Fourier Embedding) & 2 & 512 & 8.9m & 77.40 & 77.11 & 2.27 hr\\
        Transformer & 2 & 512 & 24.3m & 77.54 & 76.96 & 2.982 hr\\
        \hline
    \end{tabular}
    \end{adjustbox}
    \caption{STS-B validation scores without fine-tuning on STS-B.}
    \label{tab:model_scores}
\end{table}

With sufficient capacity (i.e., an adequately chosen \(d_{\text{model}}\) and number of layers), the Fourier-based embedding can yield representations as effective as those of learned embeddings. This finding supports the claim that semantic information can be efficiently captured by a deterministic, parameter-free mapping when combined with appropriate downstream processing. In particular, our results show that at two layers/heads with dimensions of 256 or 512, Fourier embedding converges to performance levels comparable to a standard transformer. By contrast, transformers of the same size appear over-parameterized, offering no measurable performance gains while introducing unnecessary parameters. Conversely, models quickly become under-parameterized when they use fewer than two layers/heads or a \(d_{\text{model}}\) below 256. Another advantage of Fourier embedding is its avoidance of the computational overhead associated with large embedding tables, eliminating the need for a heavy, learned matrix that grows with vocabulary size. Our custom CUDA kernel, which fuses normalization and Fourier expansion, further contributes to reduced training times. This shift effectively "frees up" parameters compared to traditional transformers, allowing for flexible downscaling or reallocation of capacity towards deeper attention layers, leading to more balanced and efficient architectures, especially when parameter counts are held constant. While the observed training time improvements are not strictly proportional to model size reductions in our small-scale experiments, they may become substantial in large language-modeling scenarios.

Ultimately, parameter size percentage reductions slow down as transformer layers are scaled horizontally (adding more layers), hence we experimented with further parameter reduction by transforming the intermediate dimensions of the MLP blocks from dim x 4 to dim / 4, equivalent to using lower rank weight matrices. Table~\ref{tab:model_comparisons} summarizes the performance of these much smaller Parameter-Efficient Transformer Embeddings (PETE) when trained with the same regime and fine-tuned on STS-B.

\begin{table}[h!]
    \centering
    \begin{adjustbox}{max width=\textwidth}
    \begin{tabular}{|c|c|c|c|c|c|c|}
        \hline
        \textbf{Model} & \textbf{Params} & \textbf{STSB Pearson-R} & \textbf{STSB Spearman-R}\\
        \hline
        PETE & 58k+ & 69.0 & 69.5\\
        \hline
        PETE & 396k+ & 78.0 & 78.0\\
        \hline
        PETE & 1.5m+ & 79.7 & 79.7\\
        \hline
        PETE & 3.6m+ & \textbf{81.7} & \textbf{81.9}\\
        \hline
        BERT-Tiny (official report) & 4m+ & 74.3 & 73.6\\
        \hline
        BERT-Mini (official report) & 11m+ & 74.3 & 73.6\\
        \hline
        TinyBERT (official report) & 14.5m & -- & 80.4\\
        \hline 
        MobileBert-Tiny (official report) & 15.1m  &  -- & 80.1\\
        \hline
        BERT-Small (official report) & 29m+ & 78.8 & 77.0\\
        \hline
    \end{tabular}
    \end{adjustbox}
    \caption{STS-B validation scores after fine-tuning on STS-B against tiny models}
    \label{tab:model_comparisons}
\end{table}
\section{Limitations}
While the results demonstrate the potential of the PETE approach, several limitations should be acknowledged. Firstly, the experiments presented here were conducted on relatively small-scale models and datasets due to resource constraints. Performance dynamics might differ significantly in larger models trained on web-scale corpora.

Secondly, the evaluation scope is currently narrow, focusing primarily on natural language inference (contrastive training) and sentence textual similarity (STS-B evaluation). The effectiveness of PETE embeddings on a broader range of NLP tasks, such as generative modeling, token-level classification (e.g., NER, POS-tagging), or complex reasoning tasks, has not yet been evaluated and remains an important area for future validation.

Thirdly, potential scalability issues may arise with extremely large vocabularies (e.g., hundreds of thousands of tokens). As the normalized token ID space \([-1, 1]\) becomes densely populated, the initial Fourier representations of distinct tokens could become very close (near-colliding embeddings). While the MLP aims to mitigate this, its ability to effectively separate a vast number of near-colliding base representations requires further investigation at scale. Furthermore, the current study does not explicitly investigate how well the deterministic Fourier expansion captures discrete lexical phenomena such as morphological variants, homonyms, or polysemy, which might be implicitly handled by traditional learned embeddings.
\section{Discussion and Conclusion}
The experimental results support our central claim: a Fourier-based, parameter-free embedding scheme can yield competitive performance compared to traditional learned embeddings while substantially reducing both the parameter count and training time. In configurations with equal overall parameters, PETE even outperforms the traditional transformer, owing to its efficient allocation of resources. By eliminating large, learned embedding tables, our method frees up capacity for deeper attention layers and better overall parameter utilization.

Future work should focus on addressing the limitations outlined above. Verifying the approach's effectiveness across different task types and at larger scales is crucial for assessing its general applicability. Investigating the interaction between the MLP and attention blocks in separating potentially near-colliding embeddings for large vocabularies will be important. Furthermore, exploring whether this embedding approach adequately captures necessary lexical nuances warrants specific study. Another promising direction is the exploration of alternative polynomial bases (e.g., Chebyshev or Legendre) for embedding mechanisms. Although these alternatives may offer theoretical advantages, our findings suggest that the Fourier basis is particularly well-suited for optimization on modern accelerators such as GPUs and TPUs. Its analytic form not only contributes to memory and computational efficiency but also opens avenues for improved interpretability of token embeddings. For example, by examining how final representations evolve as a function of token ID, one may gain insights into the model's internal semantic organization.
In conclusion, our work demonstrates that a deterministic, Fourier-based token embedding, when paired with appropriate downstream processing, can serve as an efficient alternative to learned embedding tables. This approach not only reduces the parameter burden but also reallocates capacity more effectively within the network, potentially leading to more robust and scalable models.

\end{document}